\documentclass[letterpaper, 10 pt, conference]{ieeeconf}  %

\IEEEoverridecommandlockouts                              %

\usepackage{graphics} %
\usepackage{epsfig} %
\usepackage{mathptmx} %
\usepackage{times} %
\usepackage{amsmath} %
\usepackage{amssymb}  %
\usepackage{color}
\usepackage{subfigure}
\usepackage{multicol}

\usepackage{hyperref}

\newcommand{\cmt}[1]{{\footnotesize\textcolor{red}{#1}}}

\newcommand{\todo}[1]{\cmt{TO-DO: #1}}
\newcommand{\bs}{\mathbf{s}}
\newcommand{\ba}{\mathbf{a}}
\usepackage{authblk}
\makeatletter
\renewcommand\AB@affilsepx{, \protect\Affilfont}
\makeatother

\title{\LARGE \bf
Deep Reinforcement Learning for Vision-Based Robotic Grasping:\\ A Simulated Comparative Evaluation of Off-Policy Methods
}
\DeclareRobustCommand*{\IEEEauthorrefmark}[1]{%
  \raisebox{0pt}[0pt][0pt]{\textsuperscript{\footnotesize #1}}%
}

\author[1]{Deirdre Quillen\IEEEauthorrefmark{*}\thanks{\IEEEauthorrefmark{*} Equal contribution}}
\author[1]{Eric Jang\IEEEauthorrefmark{*}}
\author[1]{Ofir Nachum\IEEEauthorrefmark{*}}
\author[2]{Chelsea Finn}
\author[1]{Julian Ibarz}
\author[1,2]{Sergey Levine}

\affil[1]{Google Brain} \affil[2]{University of California, Berkeley}

\begin{document}

\maketitle
\thispagestyle{empty}
\pagestyle{empty}

\begin{abstract}

In this paper, we explore deep reinforcement learning algorithms for vision-based robotic grasping. Model-free deep reinforcement learning (RL) has been successfully applied to a range of challenging environments, but the proliferation of algorithms makes it difficult to discern which particular approach would be best suited for a rich, diverse task like grasping. To answer this question, we propose a simulated benchmark for robotic grasping that emphasizes off-policy learning and generalization to unseen objects. Off-policy learning enables utilization of grasping data over a wide variety of objects, and diversity is important to enable the method to generalize to new objects that were not seen during training. We evaluate the benchmark tasks against a variety of Q-function estimation methods, a method previously proposed for robotic grasping with deep neural network models, and a novel approach based on a combination of Monte Carlo return estimation and an off-policy correction. Our results indicate that several simple methods provide a surprisingly strong competitor to popular algorithms such as double Q-learning, and our analysis of stability sheds light on the relative tradeoffs between the algorithms \footnote{Accompanying video: \url{https://goo.gl/pyMd6p}}.

\end{abstract} 
\section{Introduction}

Robotic grasping is one of the most fundamental robotic manipulation tasks: before interacting with objects in the world, a robot typically must begin by grasping them. Prior work in robotic manipulation has sought to address the grasping problem through a wide range of methods, from analytic grasp metrics~\cite{weisz2012pose,rodriguez2012caging} to learning-based approaches~\cite{bohg2014data}. Learning grasping directly from self-supervision offers considerable promise in this field: if a robot can become progressively better at grasping through repeated experience, perhaps it can achieve a very high degree of proficiency with minimal human involvement. Indeed, learning-based methods inspired by techniques in computer vision have achieved good results in recent years~\cite{lenz2015deep}. However, these methods typically do not reason about the sequential aspect of the grasping task, either choosing a single grasp pose~\cite{pinto2016supersizing}, or repeatedly choosing the next most promising grasp greedily~\cite{levine2016learning}. While previous works have explored deep reinforcement learning (RL) as a framework for robotic grasping in a sequential decision making context, such studies have been limited to either single objects~\cite{deepmind_grasp}, or simple geometric shapes such as cubes~\cite{tobin2017domain}.

In this work, we explore how RL can be used to automatically learn robotic grasping skills for diverse objects, with a focus on comparing a variety of RL methods in a realistic simulated benchmark. One of the most important challenges in learning-based grasping is generalization: can the system learn grasping patterns and cues that allow it to succeed at grasping new objects that were not seen during training? Successful generalization typically requires training on a large variety of objects and scenes, so as to acquire generalizeable perception and control. Prior work on supervised learning of grasping has used tens of thousands~\cite{pinto2016supersizing} to millions~\cite{levine2016learning} of grasps, with hundreds of different objects. This regime poses a major challenge for RL: if the learning is conducted primarily on-policy, the robot must repeatedly revisit previously seen objects to avoid forgetting, making it difficult to handle extremely diverse grasping scenarios. Off-policy reinforcement learning methods might therefore be preferred for tasks such as grasping, where the wide variety of previously seen objects is crucial for generalization. Indeed, the supervised learning methods explored in previous work~\cite{pinto2016supersizing,levine2016learning} can be formalized as special cases of off-policy reinforcement learning that do not consider the sequential nature of the grasping task.

Our aim in this paper is to understand which off-policy RL algorithms are best suited for vision-based robotic grasping. A number of model-free, off-policy deep reinforcement learning methods have been proposed in recent years for solving tasks such as Atari games~\cite{mnih2013playing} and control of simple simulated robots~\cite{lillicrap2015continuous}. However, these works do not explore the kinds of diverse and highly varied situations that arise in robotic grasping, and the focus is typically on final performance (e.g., expected reward), rather than generalization to new objects and situations. Furthermore, training typically involves progressively collecting more and more on-policy data, while retaining old off-policy data in a replay buffer. We study how the relative performance of these algorithms varies in an off-policy regime that emphasizes diversity and generalization.

The first contribution of this paper is a simulated grasping benchmark for a robotic arm with a two-finger parallel jaw gripper, grasping random objects from a bin. This task is available as an open-source Gym environment\footnote{Code for the grasping environment is available at \url{https://goo.gl/jAESt9}} \cite{brockman2016openai}. Next, we present an empirical evaluation of off-policy deep RL algorithms on vision-based robotic grasping tasks. These methods include the grasp success prediction approach proposed by~\cite{levine2016learning}, Q-learning~\cite{mnih2013playing}, path consistency learning (PCL)~\cite{pcl}, deep deterministic policy gradient (DDPG)~\cite{lillicrap2015continuous}, Monte Carlo policy evaluation~\cite{sutton1998reinforcement}, and Corrected Monte-Carlo, a novel off-policy algorithm that extends Monte Carlo policy evaluation for unbiased off-policy learning.

Our discussion of these methods provide a unified treatment of the various Q-function estimation techniques in the literature, including our novel proposed approach. Our results show that deep RL can successfully learn grasping of diverse objects from raw pixels, and can grasp previously unseen objects in our simulator with an average success rate of 90\%. Surprisingly, na\"{i}ve Monte Carlo evaluation is a strong baseline in this challenging domain, despite being biased in the off-policy case, and our proposed unbiased, corrected version achieves comparable performance. Deep Q-learning also excels in limited data regimes. 
We also analyze the stability of the different methods, and differences in performance across on-policy and off-policy cases and different amounts of off-policy data. Our results shed light on how the different methods compare on a realistic simulated robotic task, and suggest avenues for developing new, more effective deep RL algorithms for robotic manipulation, discussed in Section~\ref{sec:discussion}. To our knowledge, our paper is the first to provide an open benchmark for robotic grasping from image observations and held-out test objects, as well as a detailed comparison of a wide variety of deep RL methods on these tasks. 
\section{Related Work}
A number of works combine RL algorithms with deep neural network function approximators. Model-free algorithms for deep RL generally fall into one of two areas: policy gradient methods~\cite{williams1992simple,schulman2015trust,mnih2016asynchronous,wu2017scalable} and value-based methods~\cite{riedmiller2005neural,mnih2013playing,lillicrap2015continuous,gu2016continuous,haarnoja2017reinforcement}, with actor-critic algorithms combining the two classes~\cite{pcl,o2016pgq,gu2016q}. It is generally well known that model-free deep RL algorithms can be unstable and difficult to tune~\cite{islam2017reproducibility}. Most of the prior works in this field, including popular benchmarks~\cite{duan2016benchmarking,ale,brockman2016openai}, have primarily focused on applications in video games and relatively simple simulated robot locomotion tasks, and do not generally evaluate on diverse tasks that emphasize the need for generalization to new situations. The goal of this work is to evaluate which approaches are suitable for vision-based robotic grasping, in terms of both stability and generalization performance, two factors that are rarely evaluated in standard RL benchmarks.

A number of approaches have sought to apply deep RL methods for solving tasks on real robots. For example, guided policy search methods have been applied for solving a range of manipulation tasks, including contact-rich, vision-based skills~\cite{levine2016end}, non-prehensile manipulation~\cite{finn2016deep}, and tasks involving significant discontinuities~\cite{chebotar2017path,chebotar2017combining}. Other papers have directly applied model-free algorithms like fitted Q-iteration~\cite{lange12}, Monte Carlo return estimates \cite{sadeghi2017cadrl}, deep deterministic policy gradient~\cite{gu2016deep}, trust-region policy optimization~\cite{dppt}, and deep Q-networks~\cite{zhang2015towards} for learning skills on real robots. 
These papers have provided excellent examples of successful deep RL applications, but generally tackle individual skills, and do not emphasize generalizing to task instances beyond what the robot was trained on. The goal of this work is to provide a systematic comparison of deep RL approaches to robotic grasping. In particular, we test generalization to new objects in a cluttered environment where objects may be obscured and the environment dynamics are complex, in contrast to works such as \cite{tobin2017domain}, \cite{deepmind_grasp}, and \cite{james2016}, which consider grasping simple geometric shapes such as blocks.

Outside of deep RL, learning policies for grasping diverse sets of objects has been studied extensively in the literature. For a complete survey of approaches, we refer readers to Bohg et al.~\cite{bohg2014data}. Prior methods have typically relied on one of three sources of supervision: human labels~\cite{herzog2014learning,lenz2015deep}, geometric criteria for grasp success computed offline~\cite{goldfeder2009columbia}, and robot self-supervision, measuring grasp success using sensors on the robot's gripper~\cite{pinto2016supersizing}. Deep learning has been recently incorporated into such systems~\cite{kappler2015leveraging,lenz2015deep,levine2016learning,mahler2017dex,pas2017grasp}.  These prior methods do not consider the sequential decision making formalism of grasping maneuvers, whereas our focus in this paper is on evaluating RL algorithms for grasping. We do include a comparison to a prior method that learns to predict grasp outcomes without considering the sequential nature of the task~\cite{levine2016learning}, and observe that deep RL methods are more suitable in harder, more cluttered environments.

Finally, a primary consideration of this paper is the ability to effectively learn from large amounts of off-policy data, which makes deploying new algorithms much more practical. Sadheghi et al. use deep reinforcement learning to from offline simulated data to learn a model for drone flight ~\cite{sadeghi2017cadrl}. Other papers have considered large-scale data collection for robotics. For example, Finn et al. learn a predictive model of sensory inputs and used it to plan~\cite{finn2016unsupervised,finn2017deep}. Pinto \& Gupta~\cite{pinto2016supersizing} and Levine et al.~\cite{levine2016learning} both use supervised learning techniques for learning to grasp. Unlike these prior approaches, we focus on model-free RL algorithms, which can consider the future consequences of their actions (e.g., in order to enable pregrasp manipulation).

%
\section{Preliminaries}

We first define the RL problem and the notation that we use in the rest of the paper. We consider a finite-horizon, discounted Markov decision process (MDP): at each timestep $t$, the agent will observe the current state $\bs_t \in S$, take an action $\ba_t \in A$, and then receive a reward $r(\bs_t,\ba_t)$ and observe the next state $\bs_{t+1}$, each stochastically determined by the environment. Episodes have length $T$ timesteps. The goal of the agent is to find a policy $\ba\sim\pi_\theta(-|\bs)$, parameterized by $\theta$, under which the expected reward is maximized. We will additionally assume that future rewards are discounted by $\gamma$, such that the objective becomes:
\begin{equation}
\max_\theta \mathbb{E}_{\bs,\ba \sim \pi_\theta}\left[\sum_{t=1}^T \gamma^{t-1} r(\bs_t,\ba_t)\right].
\label{eq:objective}
\end{equation}
Note that the expectation is with respect to both the policy and the environment dynamics.  To reduce notational clutter, we use $\mathbb{E}$ without a subscript to refer to expectation only over the environment dynamics (and not a specific policy) and specify a specific policy only when relevant.

\section{Problem Setup}
\label{sec:problem}

\begin{figure}[h]
\centering
\centering
\includegraphics[width=0.4 \textwidth]{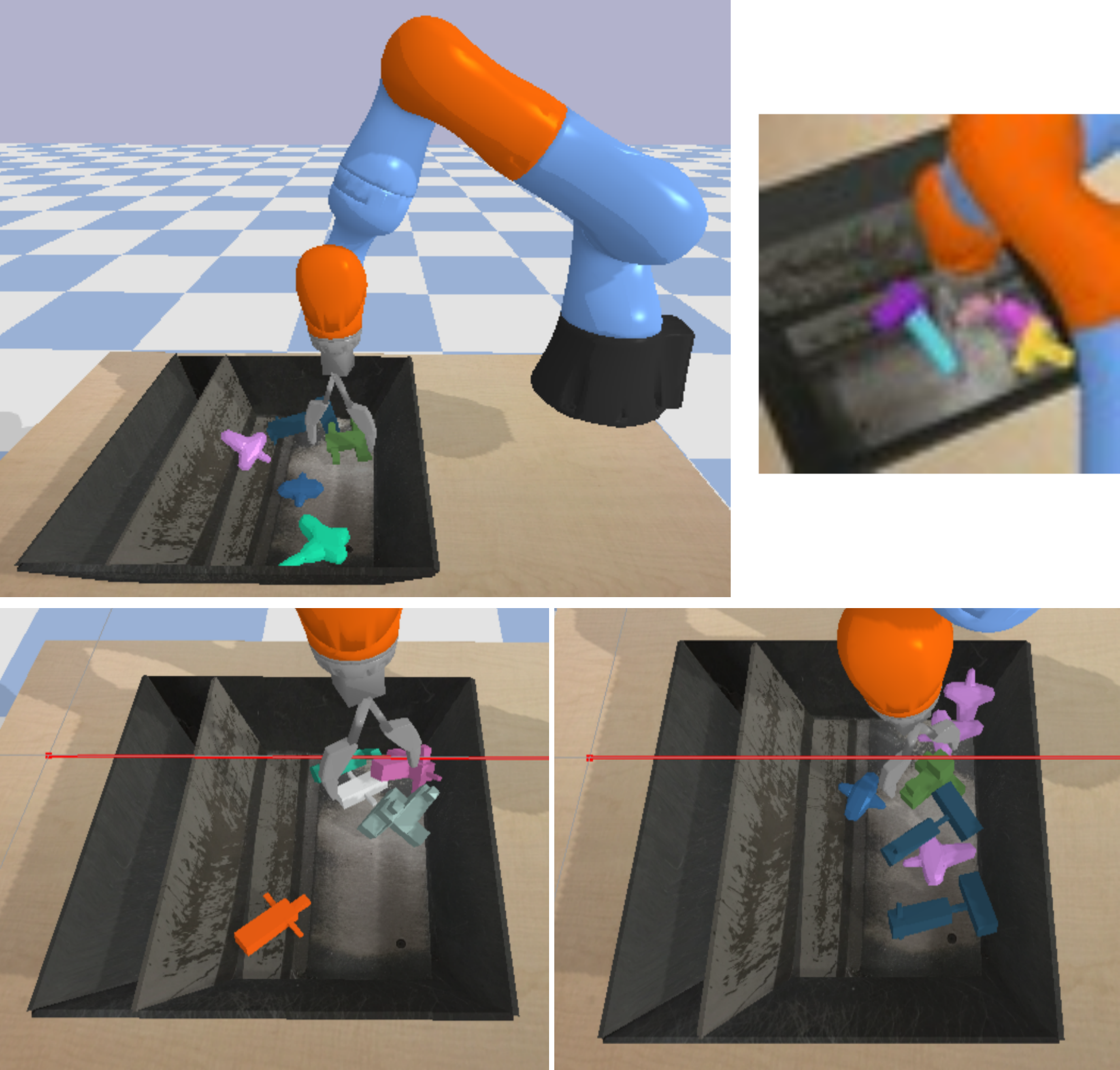}
\caption{Upper Left: An illustration of our simulated grasping setup. The robot must pick up objects in a bin, which is populated using randomized objects shown in Figure~\ref{fig:random_objects}. Upper right: Example observations to the robot. Bottom left: In the first task, the robot picks up a wide variety of randomized objects and generalizes to unseen test objects. Bottom right: In the second task, the robot has to pick one of the purple cross-shaped objects from a cluttered bin.
}
\label{fig:tasks}
\end{figure}

Our proposed benchmark for vision-based robotic grasping build on top of the Bullet simulator \cite{coumans2017}. In this environment, a robotic arm with 7 degrees of freedom attempts to grasp objects from a bin. The arm has a fixed number of timesteps ($T=15$) to find a good grasp, at which point the gripper closes and the episode ends. The reward is binary and provided only at the last step, with $r(\bs_T,\ba_T) = 1$ for a successful grasp and 0 for a failed grasp. The observed state $\bs_t$ consists of the current RGB image from the viewpoint of the robot's camera and the current timestep $t$ (Figure \ref{fig:tasks}). The timestep is included in the state, since the policy must know how many steps remain in the episode to decide whether, for example, it has time for a pre-grasp manipulation, or whether it must immediately move into a good grasping position. The arm moves via position control of the vertically-oriented gripper. Continuous actions are represented by a Cartesian displacement $[dx, dy, dz, d\phi]$, where $\phi$ is a rotation of the wrist around the $z$-axis. The gripper automatically closes when it moves below a fixed height threshold, and the episode ends. At the beginning of each new episode, the object positions and rotations are randomized within the bin.

\begin{figure}[h]
\centering
\subfigure[]{\includegraphics[width=0.23 \textwidth]{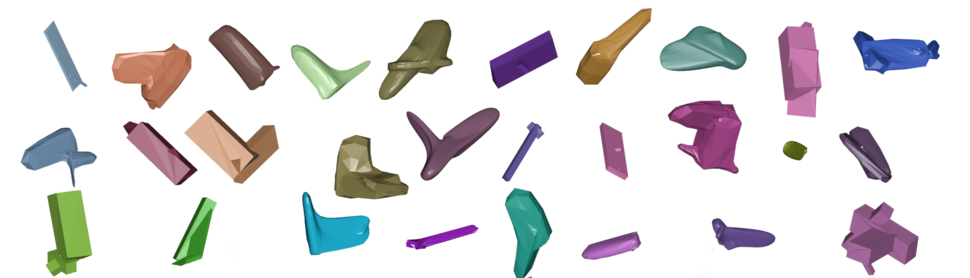}}\quad
\subfigure[]{\includegraphics[width=0.23 \textwidth]{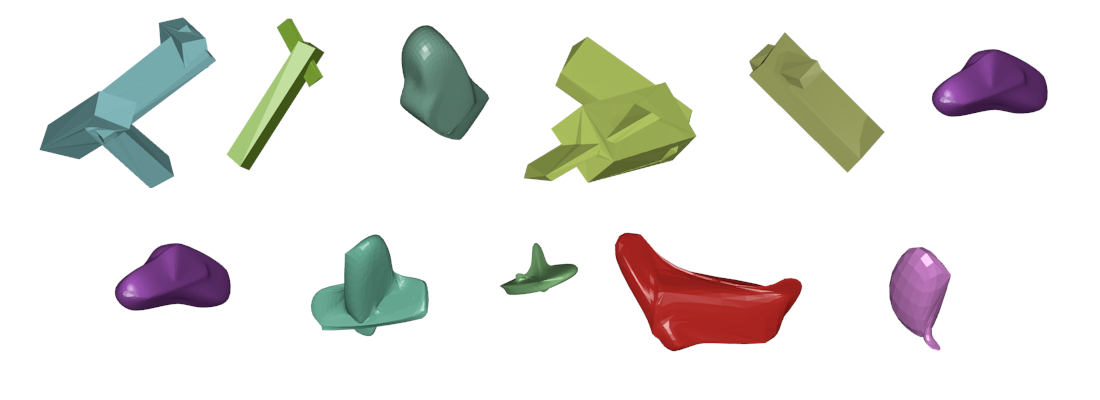}}
\caption{
\label{fig:random_objects}
Left: 30 of the 900 train objects. Right: 10 of the 100 test objects.}
\end{figure}

The benchmark consists of two different RL environments, shown in in Figure \ref{fig:tasks}.
\begin{enumerate}
\item {\bf Regular grasping.} The first grasping task tests generalization, with 900 randomly generated rigid objects with diverse random shapes used during training, and 100 testing performed on new objects on which the model was never trained previously. In each episode there are 5 objects in the bin. Every 20 episodes, the objects are randomly switched out. Both training and test objects are visualized in Figure~\ref{fig:random_objects}.

\item {\bf Targeted grasping in clutter.}
In this task, the robot must pick up a particular cross-shaped object in a bin with many other objects, which may occlude each other visually (See Figure \ref{fig:tasks} right). The arm may disturb other objects in the bin when attempting to select and grasp the ``target object''. We chose this setting because grasping specific objects in clutter may require more nuanced behavior from the robot. The robot trains on objects which are kept the same for all episodes. We evaluate performance on sets of 7 objects where 3 of them are ``target" objects, and the robot only receives reward for picking up one of the target objects. 
\end{enumerate}

%
\section{Reinforcement Learning Algorithms}

In addition to proposing a vision-based grasping benchmark, we aim to evaluate off-policy deep RL algorithms to determine which methods are best suited for learning complex robotic manipulation skills, such as grasping, in diverse settings that require generalization to novel objects. Our detailed experimental evaluation includes well-known algorithms such as Q-learning~\cite{watkins_q,ddqn}, deep deterministic policy gradient (DDPG)~\cite{lillicrap2015continuous}, which we show to be a variant of Q-learning with approximate maximization, path consistency learning (PCL)~\cite{pcl}, Monte Carlo policy evaluation~\cite{sutton1998reinforcement}, which consists of simple supervised regression onto estimated returns, and a novel corrected version of Monte Carlo policy evaluation, which makes the algorithm unbiased in the off-policy case, with a correction term that resembles Q-learning and PCL.

\subsection{Learning to Grasp with Supervised Learning}

The first method in our comparison is based on the grasping controller described by Levine et al.~\cite{levine2016learning}. This method does not consider long-horizon returns, but instead uses a greedy controller to choose the actions with the highest predicted probability of producing a successful grasp. We include this approach in our comparison because it is a recent example of a prior grasping method that learns to perform closed-loop feedback control using deep neural networks from raw monocular images. To our knowledge, no prior method learns vision-based robotic grasping with deep networks for grasping of diverse objects with reinforcement learning, making this prior approach the closest point of comparison.

This prior method learns an outcome predictor $Q_\theta(\bs,\ba)$ for the next-step reward after taking a single action $\ba$. This amounts to learning a single-step Q-function. To obtain labeled data from multi-step grasping episodes, this method uses ``synthetic'' actions obtained by taking the position of the gripper at any point during the episode, denoted $\mathbf{p}_t$, and computing the action that would move the gripper to the final pose of the episode $\mathbf{p}_T$. Since actions correspond to changes in gripper pose, the action label is simply given by $\mathbf{a}_t = \mathbf{p}_T - \mathbf{p}_t$, and the outcome of the entire episode is used as the label for each step within that episode. This introduces bias: taking a straight-line path from $\mathbf{p}_t$ to $\mathbf{p}_T$ does not always produce the same grasp outcome as the actual sequence of intermediate steps taken in the corresponding episode.
The action $\ba_t$ is selected by maximizing the Q-function $Q_\theta(\bs,\ba)$ via stochastic optimization. In our implementation, we employ the cross-entropy method (CEM), with 3 iterations and 64 samples per iteration. For further details, we refer the reader to prior work~\cite{levine2016learning}.

\subsection{Off-Policy Q-Learning}

We begin by describing the standard off-policy Q-learning algorithm~\cite{watkins_q}, which is one of the best known and most popular methods in this class. Q-learning aims to estimate the Q-function by minimizing the Bellman error, given by
\begin{equation}
\mathcal{E} = \frac{1}{2} \mathbb{E}_{\bs, \ba}\left[
\left( Q_\theta(\bs,\ba) - (r(\bs,\ba) + \gamma\max_{\ba'}Q_\theta(\bs',\ba')) \right)^2
\right]. \label{eq:bellman}
\end{equation}
Expectations over $\bs,\ba$ correspond to state-action pairs sampled from an off-policy replay buffer. Minimizing this quantity for all states results in the optimal Q-function, which induces an optimal policy. In practice, the Bellman error is minimized only at sampled states, by computing the gradient of Equation~(\ref{eq:bellman}) with respect to the Q-function parameters $\theta$ and using stochastic gradient descent. The gradient is computed only through the $Q_\theta(\bs,\ba)$ term, without considering the derivative of the non-differentiable $\max$ operator. Applying Q-learning off-policy is then straightforward: batches of states, actions, rewards, and subsequent states, of the form $(\bs_t,\ba_t,\bs_{t+1},r_t)$ are sampled from the buffer of stored transition tuples, and the gradient of the Bellman error is computed on these samples. In practice, a number of modifications to this method are employed for stability, as suggested in prior work~\cite{mnih2013playing}. First, we employ a \emph{target network} inside the $\max$ that is decorrelated from the learned Q-function, by keeping a lagged copy of the Q-function that is delayed by 50 gradient updates. We refer to this target network as $Q_{\theta'}$. Second, we employ double Q-learning (DQL)~\cite{ddqn}, which we found in practice improves the performance of this algorithm. In double Q-learning, the $\max$ operator uses the action that maximizes the current network $Q_\theta$, but the value obtained from the target network $Q_{\theta'}$, resulting in the following error estimate:
\[
\mathcal{E} \!=\! \frac{1}{2} \!\mathbb{E}_{\bs, \ba}\left[
\!\left(\! Q_\theta(\bs,\ba) \!-\! (r(\bs,\ba) \!+\! \gamma Q_{\theta'}(\bs',\arg\max_{\ba'} Q_\theta(\bs',\ba'))) \!\right)\!^2
\right].
\]
To handle continuous actions, we use a simple stochastic optimization method to compute the $\arg\max$ in the target value: we sample $16$ actions uniformly at random, and pick the one with the largest Q-value. While this method is crude, it is efficient, easy to parallelize, and we found it to work well for our 4-dimensional action parameterization. The action at execution time is selected with CEM, in the same way as described in the previous section.

\subsection{Regression with Monte Carlo Return Estimates}

Although the Q-learning algorithm discussed in the previous section is one of the most commonly used and popular Q-function learning methods for deep reinforcement learning, it is far from the simplest. In fact, if we can collect on-policy data, we can estimate Q-values directly with supervised regression. Assuming episodes of length $T$, the empirical loss for Monte Carlo policy evaluation~\cite{sutton1998reinforcement} is given by
\[
\mathcal{E} = \frac{1}{2} \sum_{i=1}^N \sum_{t=1}^T
\left( Q_\theta(\bs_t,\ba_t) - \sum_{t'=t}^T \gamma^{t'-t} r(\bs_t,\ba_t) \right)^2,
\]
where the first sum is taken over sampled episodes, and the second sum over the time steps within each episode. If the samples are drawn from the latest policy, this method provides an unbiased approach to estimating Q-values. Monte Carlo return estimates were previously used to learn deep reinforcement learning polices for drone flight in \cite{sadeghi2017cadrl}. In contrast to Q-learning, it does not require bootstrapping -- the use of the most recent function approximator or target network to estimate target values. This makes the method very simple and stable, since the optimization reduces completely to standard supervised regression. However, the requirement to obtain on-policy samples severely limits the applicability of this approach for real-world robotic manipulation. In our experiments, we evaluate how well this kind of Q-function estimator performs when employed on off-policy data.  Surprisingly, it provides a very competitive alternative, despite being a biased estimator in the off-policy case.

\subsection{Corrected Monte Carlo Evaluation}

The Monte Carlo (MC) policy evaluation algorithm described in the previous section is a well-known method for estimating Q-values~\cite{sutton1998reinforcement}, but not an especially popular one: it does not benefit from bootstrapping, and is biased when applied in an off-policy setting. We can improve this approach by removing the off-policy bias through the addition of a correction term, which we describe in this section. This correction is a novel contribution of our paper, motivated by the surprising effectiveness of the na\"{i}ve Monte Carlo evaluation method. Let $Q^*$ and $V^*$ be the Q-values and state values of the optimal policy:
\vspace{-1mm}
\begin{eqnarray}
Q^*(\bs_t, \ba_t) &=& \mathbb{E}_{\bs, \ba}[r(\bs_t, \ba_t) + \gamma V^*(\bs_{t+1})], \\
V^*(\bs_t) &=& \max_{\ba} Q^*(\bs_t, \ba).
\end{eqnarray}
\vspace{-1mm}
We may express the advantage of a state-action pair as
\vspace{-1mm}
\begin{eqnarray}
A^*(\bs_t, \ba_t) &=& Q^*(\bs_t, \ba_t) - V^*(\bs_t) \\
&=& \mathbb{E}_{\bs, \ba}[r(\bs_t, \ba_t) + \gamma V^*(\bs_{t+1}) - V^*(\bs_t)].
\end{eqnarray}
\vspace{-1mm}
Thus we have
\vspace{-1mm}
\begin{equation}
\mathbb{E}_{\bs, \ba}[V^*(\bs_t) - \gamma V^*(\bs_{t+1})] = \mathbb{E}_{\bs, \ba}[r(\bs_t, \ba_t) - A^*(\bs_t, \ba_t)].
\label{eq:consistency}
\end{equation}
If we perform a discounted sum of the two sides of Equation~\ref{eq:consistency} over $\bs_t,\dots,\bs_T$ we induce a telescoping cancellation:
\begin{multline}
\mathbb{E}_{\bs, \ba}\left[\sum_{t'=t}^T\gamma^{t'-t}(V^*(\bs_{t'}) - \gamma V^*(\bs_{t'+1}))\right] = \\ \mathbb{E}_{\bs, \ba}\left[\sum_{t'=t}^T\gamma^{t'-t} (r(\bs_{t'}, a_{t'}) - A^*(\bs_{t'}, \ba_{t'}))\right]
\end{multline}
\begin{equation}
\Rightarrow V^*(\bs_t) = \mathbb{E}_{\bs, \ba}\left[\sum_{t'=t}^T\gamma^{t'-t} (r(\bs_{t'}, \ba_{t'}) - A^*(\bs_{t'}, \ba_{t'}))\right],
\end{equation}
where we recall that $V^*(\bs_{T+1})=0$.  Equivalently, we have
\begin{equation}
Q^*(\bs_t,\ba_t) \!=\! \mathbb{E}_{\bs, \ba}\!\left[r(\bs_t, \ba_t) \!+\!\! \!\!\sum_{t'=t+1}^T\gamma^{t'-t} (r(\bs_{t'}, \ba_{t'}) \!-\! A^*(\bs_{t'}, \ba_{t'}))\right].
\label{eq:fullconsistency}
\end{equation}
Thus, we may train a parameterized $Q_\theta$ to minimize the squared difference between the LHS and RHS of Equation~\ref{eq:fullconsistency}.  Note that this resulting algorithm is a modified version of Monte Carlo augmented with a correction to the future reward given by the discounted sum of advantages. Another interpretation of this correction is the difference between the Q-values of the actions actually taken along the sampled trajectory, and the optimal actions. This means that ``good'' actions along ``bad'' trajectories are given higher values, while ``bad'' actions along ``good'' trajectories are given lower values. This removes the bias of Monte Carlo when applied to off-policy data. We also note that this corrected Monte Carlo may be understood as a variant of PCL~\cite{pcl}, discussed below, without the entropy regularization. In practice, we also multiply the correction term by a coefficient $\nu$, which we anneal from 0 to 1 during training to improve stability. When $\nu = 0$, the method corresponds to supervised regression, and when $\nu=1$, it becomes unbiased.

\subsection{Deep Deterministic Policy Gradient}

Deep deterministic policy gradient (DDPG)~\cite{lillicrap2015continuous} is an algorithm that combines elements of Q-learning and policy gradients. Originally derived from the theory of deterministic policy gradients, this algorithm aims to learn a deterministic policy $\pi_\phi(\bs) = \ba$, by propagating gradients through a critic $Q_\theta(\bs,\ba)$. However, DDPG can also be interpreted as an approximate Q-learning algorithm. To see this, observe that, in Q-learning, the policy that is used at test time is obtained by solving $\pi^*(\bs) = \arg\max_{\ba}Q_\theta(\bs,\ba)$. In continuous action spaces, performing this optimization at every decision step is computationally expensive. The actor, which is trained in DDPG according to the objective
\vspace{-1.5mm}
\begin{equation}
\max_\phi \mathbb{E}_{\bs, \ba}\left[Q_\theta(\bs,\pi_\phi(\bs))\right],\label{eq:ddpg}
\vspace{-2mm}
\end{equation}
can be seen as an approximate maximizer of the Q-function with respect to the action at any given state $\bs$. This amortizes the search over actions. The update equations in DDPG closely resemble Q-learning. The Q-function is updated according to the gradient of the bootstrapped objective
\[
\mathcal{E} \!=\! \frac{1}{2} \!\mathbb{E}_{\bs, \ba}\left[
\!\left(\! Q_\theta(\bs,\ba) \!-\! (r(\bs,\ba) \!+\! \gamma Q_{\theta'}(\bs', \pi_\phi(\bs'))) \!\right)\!^2
\right],
\]
and the actor is updated by taking one gradient step for the maximization in Equation~(\ref{eq:ddpg}). Comparing this equation to that of standard double Q-learning, we see that the only difference for the Q-function update is the use of $\pi_\phi(\bs')$ instead of the $\arg\max$. The practical implementation of DDPG closely follows that of Q-learning, with the addition of the actor update step after each Q-function update. In practice, a lagged (``target network'') version of the actor is used to compute the target value~\cite{lillicrap2015continuous}.

\subsection{Path Consistency Learning}

Path consistency learning (PCL)~\cite{pcl} is a stochastic optimal control variant of Q-learning that resembles our corrected Monte Carlo method. Though the full derivation of this algorithm is outside the scope of this paper, we briefly summarize its implementation, and include it for comparison due to its similarity with corrected MC. PCL augments the RL objective in Equation~(\ref{eq:objective}) with a $\tau$-weighted discounted entropy regularizer,
\vspace{-1mm}
\begin{equation*}
\pi^* = \arg\max_{\pi_\theta} ~\mathbb{E}_{\bs,\ba \sim \pi_\theta}\left[\sum_{t=1}^T \gamma^{t-1} (r(\bs_t,\ba_t) - \tau\log\pi_\theta(\ba_t|\bs_t))\right].
\label{eq:pclobj}
\vspace{-1mm}
\end{equation*}
The corresponding optimal value function is given by
\vspace{-1mm}
\begin{equation*}
V^*(\bs_t) =\mathbb{E}_{\bs,\ba \sim \pi^*}\left[\sum_{t'=t}^T \gamma^{t'-t} (r(\bs_{t'},\ba_{t'}) - \tau\log\pi^*(\ba_{t'}|\bs_{t'}))\right],
\end{equation*}
and together the policy and value function must satisfy $d$-step consistency for any $d>0$:
\begin{multline}
V^*(\bs_t) = \\ \mathbb{E}_{\bs, \ba}\left[\gamma^d V^*(\bs_{t+d}) + \sum_{t'=t}^{t+d-1} \gamma^{t'-t}(r(\bs_{t'}, \ba_{t'}) - \tau\log\pi^*(\ba_{t'}|\bs_{t'}))\right].
\label{eq:pcleqn}
\end{multline}
PCL minimizes the squared difference between the LHS and RHS of Equation~\ref{eq:pcleqn} for a parameterized $\pi_\theta,V_\phi$.  In our experiments, we use a variant Trust-PCL~\cite{trustpcl}, which uses a Gaussian policy and modifies the entropy regularizer to relative entropy with respect to a prior version of the policy.

\subsection{Summary and Unified View}
\label{sec:unifying}

In this section, we provide a unified view that summarizes the individual choices made in each of the above algorithms. All of the methods perform regression onto some kind of target value to estimate a Q-function, and the principal distinguishing factors among these methods consist of the following two choices:

\paragraph{Bootstrapping or Monte Carlo returns} The standard Q-learning algorithm and DDPG use the bootstrap, by employing the current function approximator $Q_\theta$ (or, in practice, a target network $Q_{\theta'}$) to determine the value of the policy at the next step, via the term $\max_{\ba'} Q(\bs',\ba')$. In contrast, both Monte Carlo variants, PCL, and the single-step supervised method use the actual return of the entire episode. This is in general biased in the off-policy case, since the current policy might perform better than the policy that collected the data. In the case of Monte Carlo with corrections and PCL, a correction term is added to compensate for the bias. We will see that adding the correction to Monte Carlo substantially improves performance.

\paragraph{Maximization via an actor, or via sampling} The DDPG and PCL methods use a second network to choose the actions, while the other algorithms only learn a Q-function, and choose actions by maximizing it with stochastic search. The use of a separate actor network has considerable benefits: obtaining the action from the actor is much faster than stochastic search, and the actor training process can have an amortizing effect that can accelerate learning~\cite{lillicrap2015continuous}. However, our empirical experiments show that this comes at a price: learning a value function and its corresponding actor function jointly makes them co-dependent on each other's output distribution, resulting in instability.

The following table summarizes the specific choices for these two parameters made by each of the algorithms:
\begin{table}[h!]
\centering
 \begin{tabular}{|l l l|} 
 \hline
 algorithm & target value & action selection \\
 \hline\hline
 supervised learning & episode value\footnotemark & stochastic search \\ 
 Q-learning & bootstrapped & stochastic search \\
 Monte Carlo & episode value & stochastic search \\
 Corrected Monte Carlo\!\!\! & corrected episode value\!\!\! & stochastic search \\
 DDPG & bootstrapped & actor network \\
 PCL & corrected episode value\footnotemark \!\!\! & actor network \\
 \hline
 \end{tabular}
\end{table}
\addtocounter{footnote}{-1}
\footnotetext{Supervised learning uses the actual episode value, but does not use the actual actions of the episode, instead merging multiple actions into a cumulative action that leads to the episode's final state.}
\addtocounter{footnote}{1}
\footnotetext{PCL also includes an entropy regularizer in the objective.}

\section{Experiments}

\begin{figure}[h]
\begin{center}
\includegraphics[width=.3\textwidth]{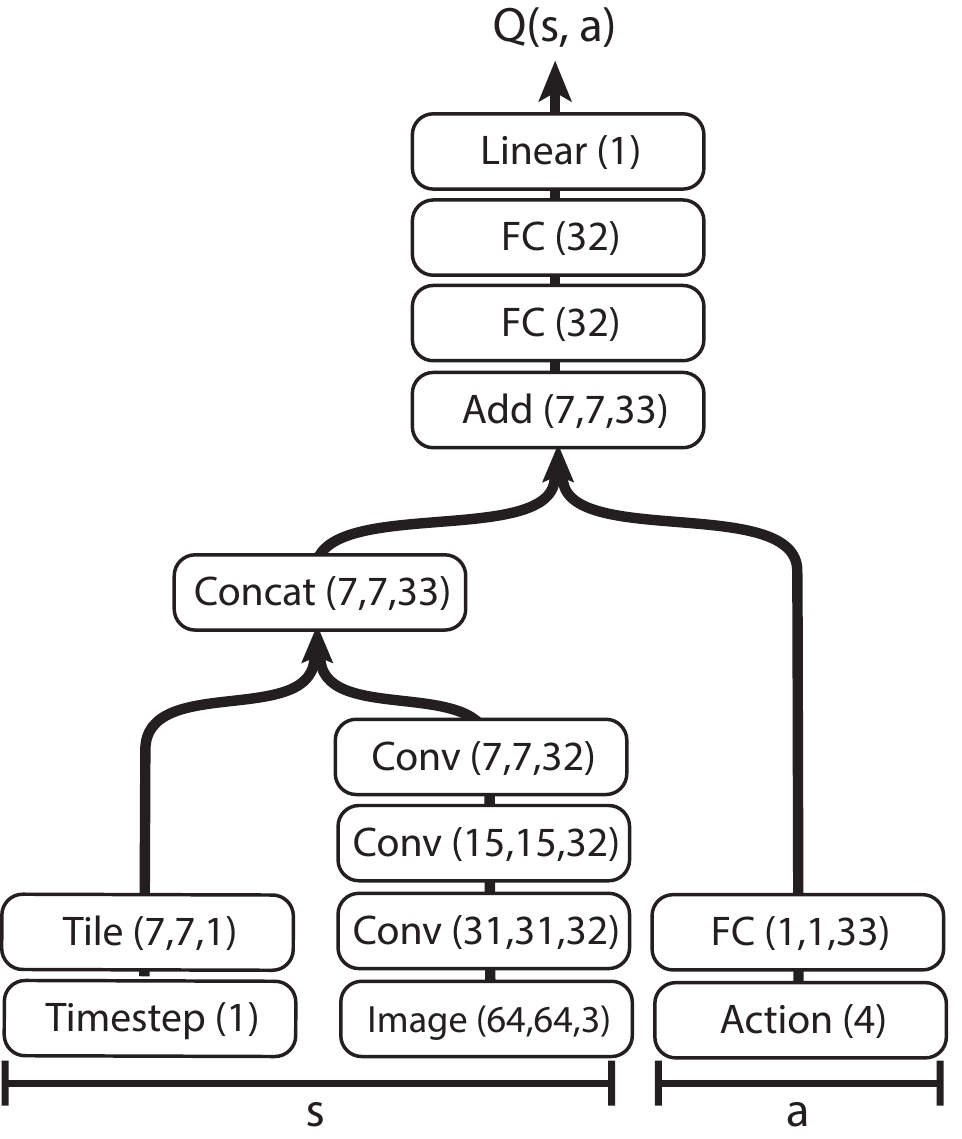}
\end{center}
\caption{
\label{fig:critic_model}
Q network architecture used for grasping tasks, with layer sizes in parentheses. The model takes as input the timestep $\textbf{t}$, the image observation $\bs$, and candidate action $\ba$.}
\end{figure}

We evaluate each RL algorithm along four axes: overall performance, data-efficiency, robustness to  off-policy data, and hyperparameter sensitivity, all of which are important for the practicality of applying these methods to real robotic systems.
As discussed in Section~\ref{sec:problem}, we consider two challenging simulated grasping scenarios, regular and targeted grasping, with performance evaluated on held-out test objects. All algorithms use variants of the deep neural network architecture shown in Figure \ref{fig:critic_model} to represent the Q-function.

\subsection{Data Efficiency and Performance}

\begin{figure}[h]
\includegraphics[width=.5\textwidth]{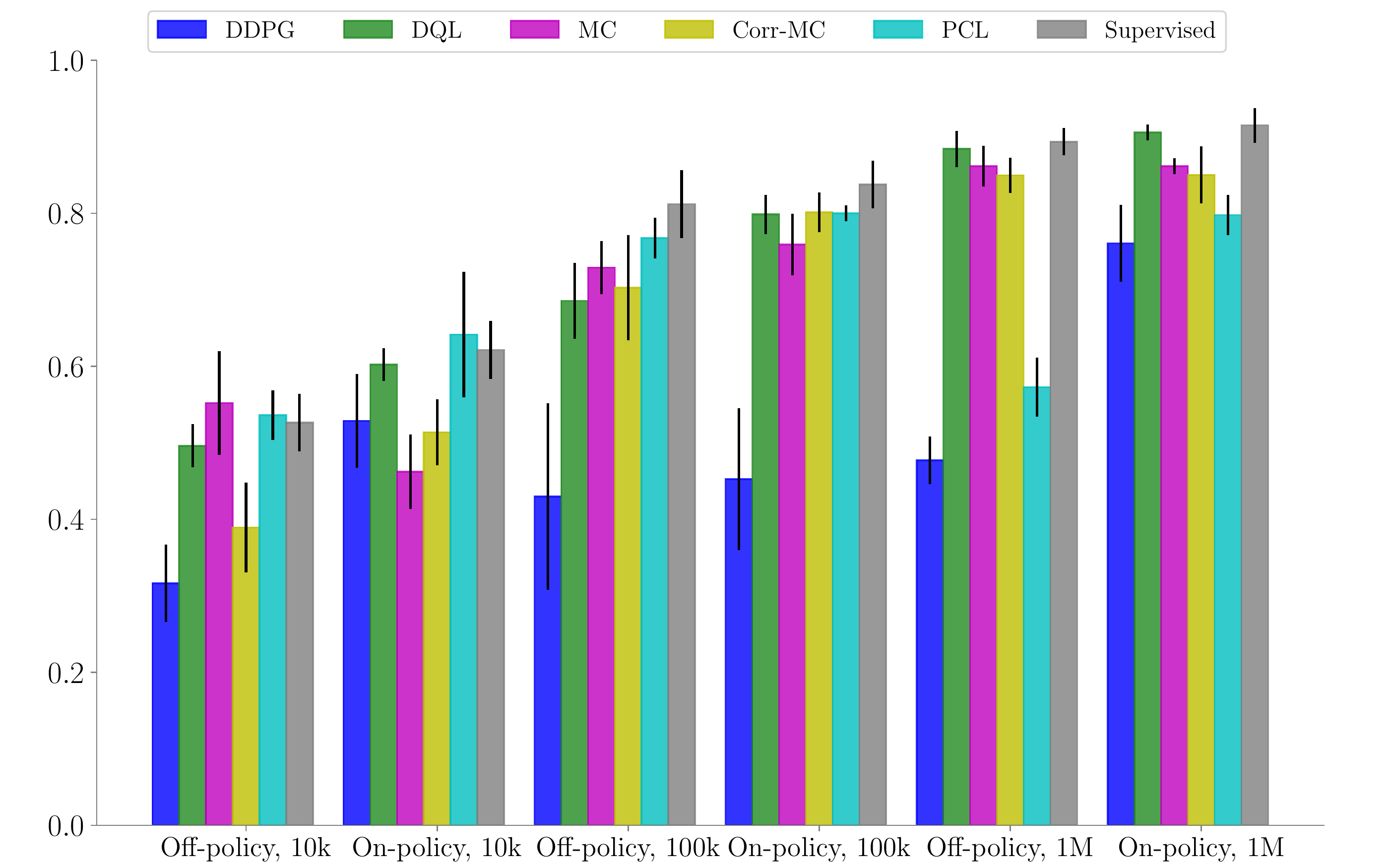}
\caption{
\label{fig:barplot1}
Regular grasping performance on held-out test objects for varying dataset sizes. DQL and the supervised baseline perform best. Standard deviations computed from 9 independent runs with different random seeds.}
\vspace{-3mm}
\end{figure}

\begin{figure}[h]
\includegraphics[width=.5\textwidth]{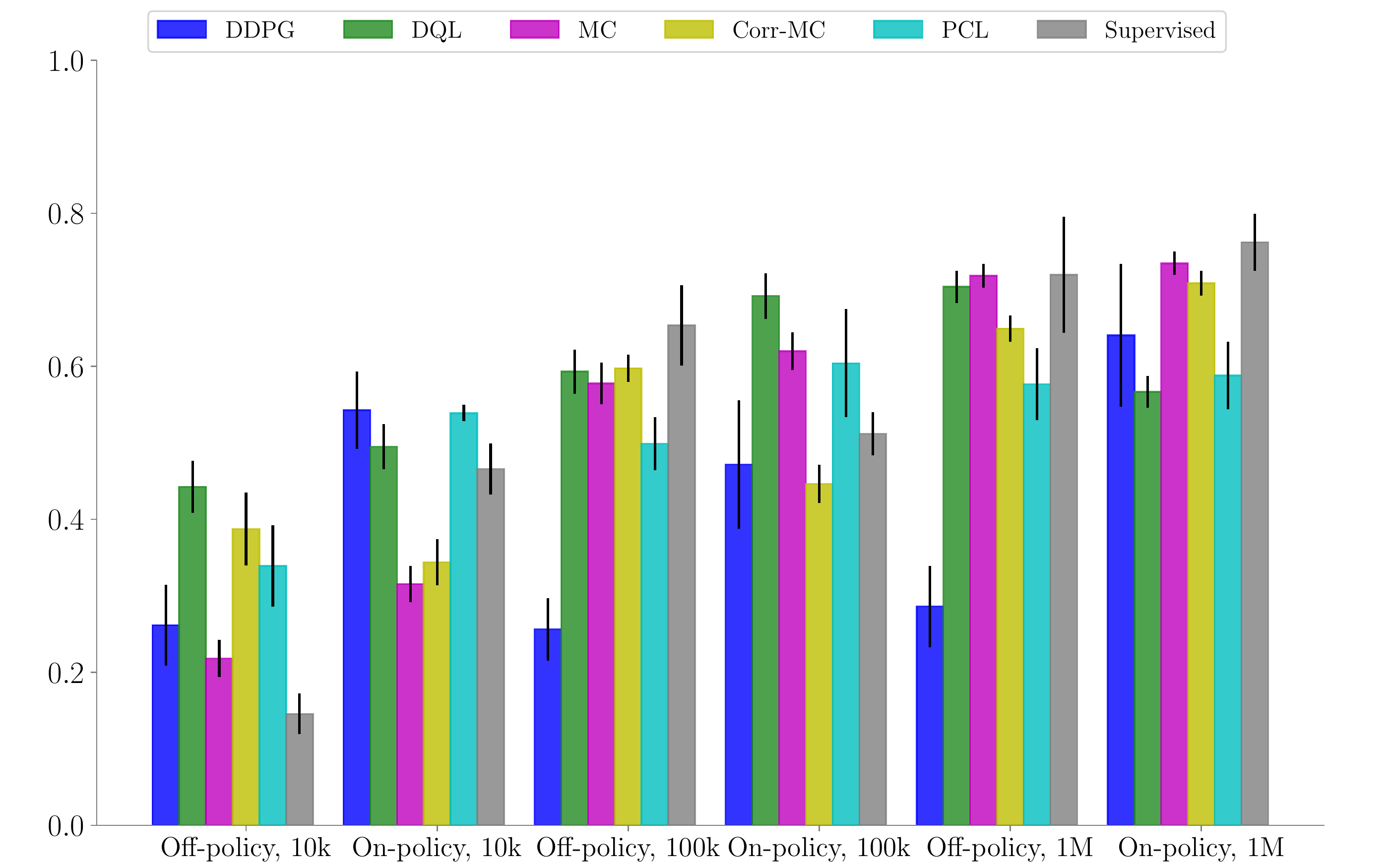}
\caption{
\label{fig:barplot3}
Targeted grasping performance in a cluttered bin with three target object and four non-target objects, for varying dataset sizes. DQL performs well in the low-data and off-policy regimes, whereas MC and corrected-MC performs best with maximal data.}
\vspace{-3mm}
\end{figure}

We consider learning with both on-policy and off-policy data. 
In each setting we initialize the pool of experience with an amount of random-policy data (10k, 100k, or 1M grasps)\footnote{Code for the random policy is available at \url{https://goo.gl/hPS6ca}}. A Q-function model is trained from this data.  In the on-policy case we periodically sample 50 on-policy grasps every 1k training steps which are used to augment the initial pool.
This setting is on-policy in the sense that we continually recollect data with the latest policy. However, the amount of on-policy data is still significantly less than traditional on-policy algorithms which sample a batch of on-policy experience for each gradient step. This procedure is thus more representative of a robotic learning setting, where data collection is much more expensive than training iterations. We find that the difference between off-policy and on-policy is slight across all algorithms in all environments (see Figures~\ref{fig:barplot1}-\ref{fig:barplot3}).
This suggests that the amount of on-policy data necessary to have a significant benefit of performance is more than what we applied, and therefore likely unfeasible for robotics.

Overall, DQL, supervised learning, MC, and our Corr-MC variant learn the most successful policies given enough data (see Figures~\ref{fig:barplot1}-\ref{fig:barplot3}). DQL tends to perform better in low-data regimes, while MC and corrected MC achieve slightly better performance in the high-data regime on the harder targeted task. The good performance of DQL in low-data regimes can be partially explained by the variance reduction effect of the bootstrapped target estimate.

Although Corr-MC and standard MC perform well, often competitively with DQL in high-data regimes, standard MC does not actually perform substantially worse than Corr-MC in most cases. Although standard MC is highly biased in the off-policy setting, it still achieves good results, except in the lowest data regimes with purely off-policy data. This suggests that the bias incurred from this approach may not be as disastrous as generally believed, which should merit further investigation in future work. It is clear that while supervised training can perform well, standard model-free deep RL methods can perform competitively and, in some cases, slightly better. Generally, DDPG and PCL perform poorly compared to the other baselines.

\vspace{-2mm}

\subsection{Analyzing Stability}

\begin{figure}[h]
\begin{center}
\includegraphics[width=.46\textwidth]{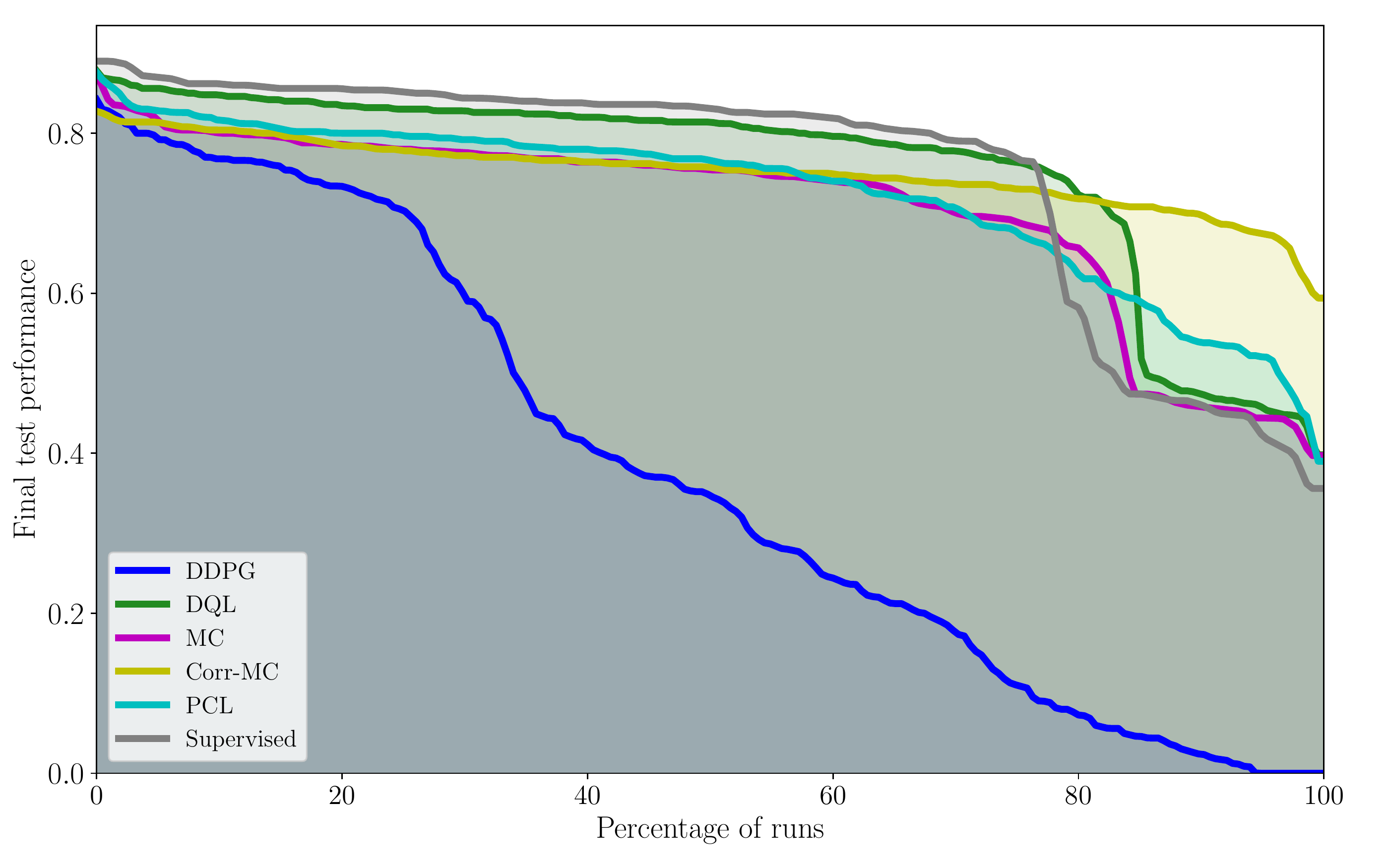}
\caption{
\label{fig:sensitivity}
Grasping success rate on held-out test objects for every hyperparameter setting in our sweep, sorted in decreasing order. DQL, PCL, and Corr-MC methods are relatively stable, while DDPG is comparatively unstable.}
\vspace{-3mm}
\end{center}
\end{figure}

When applying deep RL methods to real robotic systems, we care not only about performance and data efficiency, but robustness to different hyperparameter values. Extensively tuning a learning algorithm for a particular environment can be tedious and impractical, and current deep RL methods are known to be unstable~\cite{islam2017reproducibility}. In this section, we will study the robustness of each algorithm to hyperparameters and different random seeds. For each algorithm we sweep over different values for learning rate ($0.01$, $0.001$, $0.0001$), number of convolution filters and fully-connected units in each layer ($32$, $64$), discount factor ($0.9$, $0.95$)\footnote{MC and Supervised do not use the discount factor hyperparameter.}, and duration (in training steps) of per-step exploration with a linearly decaying schedule ($10000$, $10000$). All hyperparameter sweeps were done in the on-policy learning setting with 100k initial random grasps.

In Figure~\ref{fig:sensitivity}, we show an analysis of the sensitivity of each algorithm to each combination of the aforementioned hyperparameter values and 9 random seeds. Our results show that DQL, Corr-MC, PCL, MC, and Supervised are relatively stable across different hyperparameter values. This plot is insightful, showing that although MC and Corr-MC yield similar performance given optimal hyperparameters, the unbiased Corr-MC is slightly more robust to hyperparameter choice. The performance of DDPG drops substantially for suboptimal hyperparameters. Correspondingly, DDPG (which is the least stable) typically achieves the worst performance in our experiments. These results strongly indicate that algorithms that employ a second network for the actor suffer a considerable drop in stability, while approximate maximization via stochastic search, though crude, provides significant benefits in this regard.
\section{Discussion and Future Work}
\label{sec:discussion}

We presented an empirical evaluation of a range of off-policy, model-free deep reinforcement learning algorithms. Our set of algorithms includes popular model-free methods such as double Q-learning, DDPG, and PCL, as well as a prior method based on supervised learning with synthetic actions~\cite{levine2016learning}. We also include a na\"{i}ve Monte Carlo method, which is biased in the off-policy case but surprisingly achieves reasonable performance, often outperforming DDPG, and a corrected version of this Monte Carlo method, which is a novel contribution of this work. Our experiments are conducted in a diverse grasping simulator on two types of tasks: a grasping task that evaluates generalization to novel random objects not seen during training, and a targeted grasping task that requires isolating and grasping a particular type of object in clutter.

Our evaluation indicates that DQL performs better on both grasping tasks than other algorithms in low-data regimes, for both off-policy and on-policy learning, and additionally having the desirable property of being relatively robust to choice of hyperparameters.
When data is more plentiful, algorithms that regress to a multistep return, such as Monte Carlo or the corrected variant of Monte Carlo typically achieve slightly better performance. When considering the algorithm features summarized in Section~\ref{sec:unifying}, we find that the use of an actor network substantially reduces stability, leading to poor performance and severe hyperparameter sensitivity. Methods that use entire episode values for supervision tend to perform somewhat better when data is plentiful, while the bootstrapped DQL method performs substantially better in low data regimes. These insights suggest that, in robotic settings where off-policy data is available, single-network methods may be preferred for stability, and methods that use (corrected) full episode returns should be preferred when data is plentiful, while bootstrapped methods are better in low data regimes. A natural implication of this result is that future research into robotic reinforcement learning algorithms might focus on combining the best of bootstrapping and multistep returns, by adjusting the type of target value based on data availability. Another natural extension of our work is to evaluate a similar range of methods in real-world settings. Since the algorithms we evaluate all operate successfully in off-policy regimes, they are likely to be reasonably practical to use in realistic settings.

\section{Acknowledgements}

We thank Laura Downs, Erwin Coumans, Ethan Holly, John-Michael Burke, and Peter Pastor for helping with experiments. 
\bibliography{refs}
\bibliographystyle{abbrv}

\end{document}